\documentclass[a4paper]{article}

\usepackage{INTERSPEECH_v2}

\title{Prosodic Event Recognition using Convolutional Neural Networks with Context Information}
\name{Sabrina Stehwien, Ngoc Thang Vu}
\address{University of Stuttgart, Germany}
\email{\{sabrina.stehwien,thang.vu\}@ims.uni-stuttgart.de}

\begin{document}

\maketitle
\begin{abstract}
This paper demonstrates the potential of convolutional neural networks (CNN) for detecting and classifying prosodic events on words, specifically pitch accents and phrase boundary tones, from frame-based acoustic features. Typical approaches use not only feature representations of the word in question but also its surrounding context.
We show that adding position features indicating the current word benefits the CNN.
In addition, this paper discusses the generalization from a speaker-dependent modelling approach to a speaker-independent setup.
The proposed method is simple and efficient and yields strong results not only in speaker-dependent but also speaker-independent cases.
\end{abstract}
\noindent\textbf{Index Terms}: prosodic analysis, convolutional neural networks

\section{Introduction}

Prosodic Event Recognition (PER) refers to the task of automatically localizing pitch accents and phrase boundary tones in speech data and often deals with labelling specific segments, such as words or syllables. 
PER is important for for the analysis of human discourse and speech due to the interaction between prosody and meaning in languages such as English. For example, knowing what word in an utterance is pitch accented provides important insight into discourse structure such as focus, givenness and contrast \cite{Hirschberg1986,Selkirk1995}. Phrasing information and boundary tones for example relate to the syntactic structure \cite{Truckenbrodt99}.

A substantial amount of research has dealt with the impact of prosodic information for a wide range of language understanding tasks such as automatic speech recognition \cite{Waibel1988,Vicsi2010,Shri2007,Chen2006} and understanding \cite{Kompe,Shriberg2004,Batliner2001}. Furthermore, since manual prosodic annotation is expensive, it is desirable to have reliable, automatic annotation methods to aid linguistic and speech processing research on a large scale.

Most PER methods consist of two stages: feature extraction and preprocessing, and statistical modelling or classification. PER distinguishes two subtasks: detection typically refers to the binary classification task (presence or absence of a prosodic event), while prosodic event classification encompasses the full multi-class labelling of prosodic event types \cite{Rosenberg2007} e.g. as described in the ToBI standard \cite{Silverman}. Typically the recognition of pitch accents is modelled separately from phrase boundaries, although the acoustic features are quite similar \cite{Schweitzer2009,Rosenberg2015,Chen2004}.

Many approaches focus on finding appropriate acoustic representations of prosody \cite{Schweitzer2009,Rosenberg2007}. These features generally describe the fundamental frequency (f0) and energy and can be either frame-based \cite{Taylor1995} or grouped across segments \cite{Rosenberg2009}. Often acoustic-prosodic features also include the duration of certain segments \cite{Schweitzer2009, Tamburini2003, Sun2002}.
Most successful methods that rely on acoustic features also benefit from the addition of lexico-syntactic information \cite{Anantha2008,Schweitzer2009,Sun2002}.
Since prosodic events usually span several segments, many cited approaches add features representing the surrounding segment, while others explicitly focus on context modelling \cite{Levow2005,Rosenberg2015,Zhao2013}.

Recent work has shown that convolutional neural networks (CNN) are suitable for the detection of prominence: Shahin et al. \cite{Shahin2016} combine the output of a CNN that learns high-level features representations from 27 frame-based Mel-spectral features with global (or aggregated) f0, energy and duration features across syllables for lexical stress detection. Wang et al. \cite{Wang2016} train a CNN on continuous wavelet transformations of the fundamental frequency for the detection of pitch accents and phrase boundaries in a speaker-dependent task.

As previously pointed out in \cite{Sun2002,Rosenberg2009}, the large number of different approaches and task descriptions renders the comparison of PER performance methods quite difficult. Thus, our results are compared only to approaches that use the Boston University Radio News Corpus (BURNC) \cite{Ostendorf1995} and purely acoustic features. Some selected work with similar focus is listed in the following. 
Good results for pitch accent detection were reported by Sun \cite{Sun2002}, namely 84.7\% on one speaker (f2b) of BURNC using acoustic features only. Wang et al. \cite{Wang2016} use CNNs to detect pitch accents and phrase boundaries on the f2b speaker, obtaining 86.9\% and 89.5\% accuracy respectively.
Ren et al. \cite{Ren2004} obtain 83.6\% accuracy in speaker-independent pitch accent detection on two female speakers in BURNC.
The more difficult task is prosodic event type classification. Rosenberg \cite{Rosenberg2010} reports almost 64\% accuracy for pitch accents and 72.9\% for phrase boundaries in experiments that aimed at classifying 5 ToBI types each in 10-fold cross-validation experiments.
Chen et al. \cite{Chen2004} apply their neural-based method to speaker-independent setups using 4 speakers of BURNC and distinguishing 4 event types. They report 68.2\% recognition accuracy using only acoustic-prosodic features.
An early example of a neural network approach was proposed in \cite{Taylor1995}, and relied only on frame-based acoustic features such as f0 and energy.

In this work, we use a CNN that learns high-level feature representations on its own from low-level acoustic descriptors. This way we can rely only on frame-based features that are readily obtained from the speech signal. The only segmental information used in this work is the time-alignment at the word level. We address the notion of explicit context modelling with CNNs in a simple and efficient way.
We apply this method to both the detection and classification of pitch accents and intonational phrase boundaries.  
An additional challenge to PER is the generalization across different speakers due to the large variation in prosodic parameters. For this reason, we not only test the performance of the proposed method on one speaker for comparability, but also as leave-one-speaker-out cross-validation results. 
We report recognition accuracies comparable to similar previous work and show that our model generalizes well across speakers.

\section{Model}
\begin{figure}
\includegraphics[width=.45\textwidth]{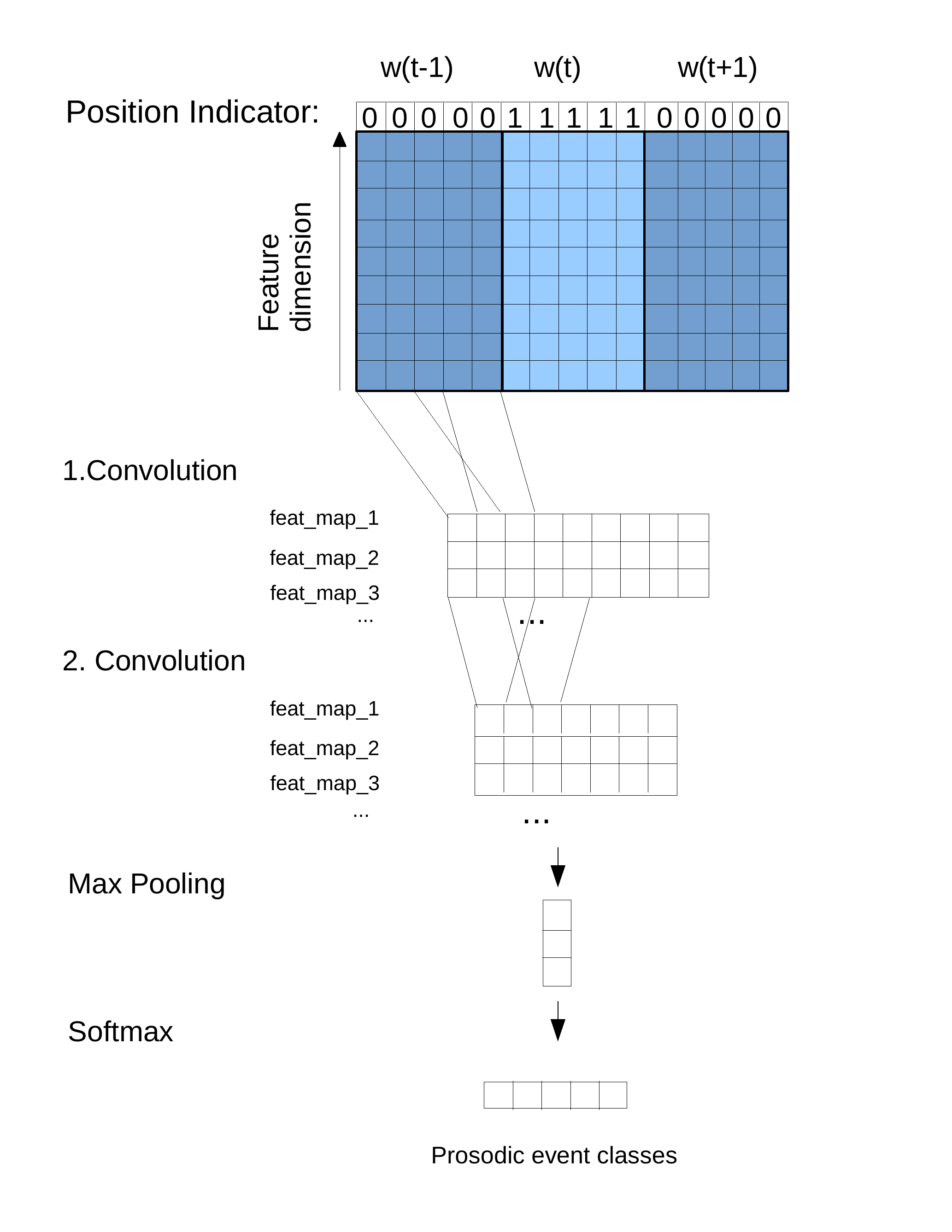}
\caption{CNN for prosodic event recognition with an input window of 3 successive words and position indicating features.}
\label{fig:model}
\end{figure}
We apply a CNN model as illustrated in Figure~\ref{fig:model} for PER. 
The task is set up as a supervised learning task in which each word is labelled as carrying a prosodic event or not.
The input to the CNN is a feature representation of the audio signal of the current word and (optionally) its context.
The signal is divided into $s$ overlapping frames and represented by a $d$-dimensional feature vector for each frame. 
Thus, for each utterance, a matrix $W \in R^{d \times s}$ is formed as input. 
The number of frames $s$ depends on the duration (signal length) of the word as well as the context window size and the frame shift.
For the convolution operation we use 2D kernels $K$ (with width $|K|$) spanning all $d$ features. The following equation expresses the convolution:
\begin{equation}
(W * K)(x,y) = \sum\limits_{i=1}^d \sum\limits_{j=1}^{|K|} W(i,j) \cdot K(x-i,y-j)
\end{equation}
We apply two convolution layers in order to expand the input information.
After the convolution, max pooling is used to find the most salient features. 
All resulting feature maps are concatenated to one feature vector which is fed into the softmax layer. The softmax layer has either 2 units for binary classification or $c$ classes for multi-class classification. 
For regularization, we also apply dropout \cite{Srivastava14} to this last layer.

\subsection{Acoustic Features}
The features used in this work were chosen to be simple and fast to obtain.
We extract acoustic features from the speech signal using the OpenSMILE toolkit \cite{opensmile}.
In this work, two different feature sets are used: a \textit{prosody} feature set consisting of 5 features from the OpenSMILE catalogue (smoothed f0, RMS energy, PCM loudness, voicing probability and Harmonics-to-Noise Ratio), and a \textit{Mel} feature set consisting of 27 features extracted from the Mel-frequency spectrum (similar to \cite{Shahin2016}). The features are computed for each 20ms frame with a 10ms shift. These two features sets are used both separately and jointly (concatenated) in the reported experiments.
The time intervals that indicate the word boundaries provided in the corpus are used to create the input feature matrices by grouping all frames for each word into one input matrix.
Afterwards, zero padding is added to ensure that all matrices have the same size.

\subsection{Position Indicator Feature}
The following describes the extension of the acoustic features by a position indicator for PER.
This type of feature has been proposed for use in neural network models for relation classification \cite{VuRelation,Zhangarxiv}.
Previous work has demonstrated the benefits of adding context information to PER \cite{Rosenberg2015,Levow2005}. The most straighforward approach is to add features that represent the right and left neighbouring segments to form a type of acoustic context window \cite{Rosenberg2007,Schweitzer2009,Wang2016}.
The caveat of using context windows as input to our CNN model is, however, that it also adds a substantial amount of noise. The learning method of CNNs is to look for patterns in the whole input and learn abstract global representations of these. The neighbouring words may have prosodic events or other prosodic prominence characteristics that ``distract'' from the current word. This effect may be amplified by the fact that the words have variable lengths. For this reason we add position features (or indicators) that are appended as an extra feature to the input matrices (see Figure~\ref{fig:model}). These features indicate the parts of the matrix that represent the current word. The rest of the matrix consists of zeros in this dimension. In the first convolution layer we ensure that the kernels always span the position-indicating feature dimension. Thus, the model is constantly informed whether the $|K|$ current frames belong to the current word or the neighbouring words. 

\section{Experimental Setup}

\subsection{Data}

The dataset used in this work is a subset of BURNC that has been manually labelled with prosodic events according to the ToBI labelling standard \cite{Silverman}. The speech data was recorded from 3 female and 2 male speakers, adding up to around 2 hours and 45 minutes of speech. Table~\ref{tab:boston} shows the number of words for each speaker in the datasets used for pitch accent and phrase boundary recognition in this work\footnote{Since the two tasks are trained and tested separately, we judge the mismatch in the two datasets as inconsequential to our experiments.}.

\begin{table}[th]
	\caption{Number of words in each subset of BURNC used in this work for pitch accent (PA) recognition and phrase boundary (PB) recognition.}
  \label{tab:boston}
  \centering
  \begin{tabular}{l||l|l|l|l|l}
    \toprule
	Speakers & f1a & f2b & f3a & m1a & m2b \\
	\hline
	PA \# words & 4375 & 12357 & 2736 & 3584 & 3607 \\
	  PB \# words & 4362 & 12606 & 2736 & 5055 & 3607 \\
    \bottomrule
  \end{tabular}  
\end{table}

For the speaker-dependent experiments, the largest speaker subset f2b is used in line with previous methods \cite{Sun2002, Wang2016}. We test our models using 10-fold cross-validation and validated on 1000 words from the respective training set.
In the speaker-independent case, the models were trained and tested using leave-one-speaker-out cross-validation and validated on 500 words from a speaker of the same gender for early stopping\footnote{This way we avoid a too large mismatch between the validation and test data.}.
All experiments are repeated 3 times and the results are averaged.

The Boston corpus contains different ToBI types of pitch accents and phrase boundaries. For the binary classification task (detection) all labels are grouped together as one class. For the classification task, we distinguish 5 different ToBI types of pitch accents and phrase boundaries (as in \cite{Rosenberg2010}), where the downstepped accents are collapsed into the non-downstepped ones: The pitch accent classes are (1) H* and !H*, (2) L*, (3) L+H* and L+!H*, (4) L*+H and L*+!H and (5) H+!H*. The boundary tones considered in this work mark the boundaries of intonational phrases: L-L\%, L-H\%, H-L\%, !H-L\%, !H-L\% and H-H\%. Uncertain events, where the annotator was unsure if there is an accent or boundary tone, are ignored for both detection and classification. Uncertain types, where the annotator was unsure of the event type, are ignored for classification.

\subsection{Hyperparameters}
The classification model is a 2-layer CNN. The first layer consists of 100 2-dimensional kernels of the shape $6 \times d$ and a stride of $4 \times 1$, with $d$ as the number of features. The kernels encompass the whole feature set to ensure that all features are learnt simultaneously. The second layer consists of 100 kernels of the shape $4 \times 1$ and a stride of $2 \times 1$. The max pooling size is set so that the output of each max pooling on each of the 100 feature maps has the shape $x$. Thus, this hyperparameter varies depending on the dimensions of the input matrix, but is kept constant due to the zero padding in each individual experiment.
Dropout with $p=0.2$ is applied before the softmax layer.
The models are trained for 50 epochs with an adaptive learning rate (Adam \cite{Adam17}) and L2 regularization.

\section{Results}

We report results for each experiment with three context variations: no context (1 word), right and left context words (3 words) and right and left context words with position features (3 words + PF).

\subsection{Pitch Accent Recognition}

\begin{table}[t]
\caption{Results (accuracy) for pitch accent recognition on speaker f2b with 10-fold cross-validation. The majority class baseline for detection is 52.1\%, for classification 48.2\%.}
  \label{tab:pa-f2b}
  \centering
  \begin{tabular}{l||l|l|l}
    \toprule
	Feature set & prosody & Mel & prosody + Mel \\
	\hline
	  \textbf{Detection} & & \\
	1 word & 84.2 & 84.2 & 84.0 \\
	3 words & 58.3 &  53.1 & 53.6  \\
	  3 words + PF & \textbf{86.3} & 83.3 & 83.9 \\
	\hline
	  \textbf{Classification} & & \\
	  1 word & 74.4 & 72.7 & 73.5 \\
	3 words & 52.4 & 47.8  & 47.8 \\
	  3 words + PF & \textbf{76.3} & 72.3 & 72.9 \\
    \bottomrule
  \end{tabular}  
\end{table}

\begin{table}[t]
\caption{Results (accuracy) for pitch accent recognition with leave-one-speaker-out cross-validation. The majority class baseline for detection is 51.5\% accuracy, for classification 48.8\%.}
  \label{tab:pa-all}
  \centering
  \begin{tabular}{l||l|l|l}
    \toprule
	Feature set & prosody & Mel & prosody + Mel \\
	\hline
	\textbf{Detection} & & \\
	1 word & 81.9 & 78.3 & 79.3 \\
	3 words & 58.2 & 54.3 & 55.3 \\
	3 words + PF & \textbf{83.6} & 80.3 & 81.1 \\
	\hline
	\textbf{Classification} & & \\
	1 word & 68.0 & 64.7 & 64.5  \\
	3 words & 50.5 & 48.4 & 48.4 \\
	3 words + PF & \textbf{69.0} & 65.9 & 65.3 \\
    \bottomrule
  \end{tabular}  
\end{table}

\begin{table}[t]
\caption{Pitch accent recognition accuracies for each speaker using prosody and position features.}
  \label{tab:pa-speakers}
  \centering
  \begin{tabular}{l||l|l|l|l|l}
    \toprule
	Speaker  & f1a & f2b & f3a & m1b & m2b \\
	\hline
	detection & 85.6  & 82.9 & 83.5 & 81.4 & 84.8  \\
	classification & 70.6 & 71.8 & 67.7 & 68.4 & 66.6  \\
    \bottomrule
  \end{tabular}  
\end{table}

Table~\ref{tab:pa-f2b} shows the results for pitch accent recognition on the single-speaker dataset and Table~\ref{tab:pa-all} shows the results obtained in speaker-independent experiments. 
The model yields up to 84\% detection performance when considering only the current word with no additional context in the speaker-dependent setup and almost 82\% in the speaker-independent experiments. The classification task is more difficult, especially in the speaker-independent case (68\%).
The results show a large drop in performance, down to the majority class baseline level, when extending the input to include the right and left context words. After adding the position indicating features, the accuracies of all tasks increases and exceeds those obtained from the single-word input in the speaker-independent case.
We obtain up to 86.3\% accuracy in pitch accent detection on f2b, which is comparable to the best previously reported results on purely acoustic input. This indicates that not only is the position indicator crucial when adding context to our specific model, but that it constitutes a strong modelling technique. Speaker-independent pitch accent classification remains the most difficult task, although the accuracy obtained in this work (69\%) matches up to that of comparable methods. 

We observe that in both the speaker-dependent and speaker-independent settings, the prosody feature set performs best, while the Mel and combined prosody + Mel feature yield similar results.

We also report the accuracies per speaker for the speaker-independent experiments using the prosody feature set and the position indicator features in Table~\ref{tab:pa-speakers}. The results show that even though the speaker f2b constitutes the largest speaker subset leaving the least amount of data for training, the model does not perform much worse than on data from other speakers. Overall, there does not appear to be a distinctively ``easy'' or ``difficult'' speaker.

\subsection{Phrase Boundaries}

\begin{table}[t]
\caption{Results (accuracy) for phrase boundary tone recognition on speaker f2b with 10-fold cross-validation. The majority class baseline for both tasks is 77.9\% accuracy.}
  \label{tab:pb-f2b}
  \centering
  \begin{tabular}{l||l|l|l}
    \toprule
	Feature set & prosody & Mel & prosody + Mel \\
	\hline
	  \textbf{Detection} & & & \\
	1 word & 87.6 & 89.2 & 89.8 \\
	3 words & 80.3 & 75.4 & 75.4 \\
	  3 words + PF & 90.2  & 90.4  & \textbf{90.5}  \\
	\hline
	  \textbf{Classification} & & & \\
	1 word & 85.6 & 87.6 & 88.0 \\
	3 words & 79.7 & 74.5 & 74.6 \\
	  3 words + PF & 87.8 & 88.7 & \textbf{88.8} \\
    \bottomrule
  \end{tabular}  
\end{table}

\begin{table}[t]
	\caption{Results (accuracy) for phrase boundary tone recognition with leave-one-speaker-out cross-validation. The majority class baseline for both tasks is 80.7\% accuracy.}
  \label{tab:pb-all-bin}
  \centering
  \begin{tabular}{l||l|l|l}
    \toprule
	Feature set & prosody & Mel & prosody + Mel \\
	\hline
	\textbf{Detection} & & & \\
	1 word & 86.5 & 85.3 & 86.1 \\
	3 words & 82.7 & 81.0 & 80.8 \\
	3 words + PF & \textbf{89.8} & 88.3 & 88.8 \\
	\hline
	\textbf{Classification} & & & \\
	1 word & 85.1 & 84.4 & 84.9 \\
	3 words & 82.5 & 81.4 & 81.5 \\
	3 words + PF & \textbf{87.3} & 86.2 & 86.7 \\
    \bottomrule
  \end{tabular}  
\end{table}

\begin{table}[t]
  \caption{Phrase boundary recognition accuracies for each speaker using prosody and position features.}
  \label{tab:pb-speakers}
  \centering
  \begin{tabular}{l||l|l|l|l|l}
    \toprule
	Speaker & f1a & f2b & f3a & m1b & m2b \\
	\hline
	detection & 88.4 & 88.8 & 91.1 & 91.4 & 89.3 \\
	classification & 86.0 & 86.1 & 87.7 & 89.0 & 87.6 \\
    \bottomrule
  \end{tabular}  
\end{table}

The results for phrase boundary recognition appear to follow a similar pattern as for pitch accent recognition. In this task, we also observe a drop in performance when extending from the 1-word to the 3-word input windows, although this effect is not as pronounced in the case of phrase boundaries. Adding position indicator features improves the results in all cases.

For the speaker-dependent task, the combined prosody and Mel feature set yields the best performance, while the small prosody feature set appears to be the best choice in the speaker-independent task. These differences, however, are not as pronounced as in the case of pitch accents.
In the f2b experiments we obtain 90.5\% and 88.8\% accuracy for detection and classification, respectively and in the speaker-independent setup we obtain almost 90\% accuracy for detection and 87.3\% for classification.

In contrast to the pitch accent recognition results, we observe that the accuracies are lowest on speaker f1a, and highest on speaker m1b in both tasks (see Table~\ref{tab:pb-speakers}).


\subsection{Discussion}

An interesting result in the above work is the impact of adding context frames without position features on the two presented tasks. We observe that adding ``uninformed'' context information is more detrimental to the recognition of pitch accents than to phrase boundaries. While we have not further examined this effect in the present study, it may be explained as follows. Pitch accents are rather local phenomena occurring on stressed syllables and are more frequent in the data. Intonational phrase boundary tones as described by the ToBI standard\footnote{\texttt{http://www.speech.cs.cmu.edu/tobi/ToBI.0.html}} not only span longer stretches of speech (since these consist of an intermediate phrase accent and an intonational phrase boundary tone) but are also more sparse since they only occur at the end of intonational phrases. This means that the model may be less sensitive to local events or changes in neighbouring segments and that it is less likely for phrase boundaries to occur in two succeeding words as in the case of pitch accents. 

\begin{table}[th!]
  \caption{Effects of z-scoring in speaker-independent experiments using prosody and position features.}
  \label{tab:pb-all-bin}
  \centering
  \begin{tabular}{l||l|l}
    \toprule
	  & non-normalized & normalized \\
	\hline
	  \textbf{Pitch Accents} & & \\ 
	  Detection & 83.6 & 77.0 \\
	 Classification & 69.0 & 62.6 \\
	\hline
	  \textbf{Phrase Boundaries} & & \\
	Detection & 89.8 & 83.0  \\
	 Classification & 87.3 & 83.2 \\
    \bottomrule
  \end{tabular}  
\end{table}

The effect of using the various feature sets in our experiments shows that the smallest feature set (prosody) works best in almost all cases, with speaker-dependent phrase boundary recognition as the only exception. These differences, however, are small. The features used in this work were chosen to be quite simple, leaving room for further investigation with respect to the acoustic features on the individual tasks.

A widely-used measure to enable the generalization of prosodic models across speakers is speaker normalization in the form of z-scoring \cite{Rosenberg2007,Chen2004,KSchweitzer2010}.
In our experiments we observe a large drop in performance after z-scoring the features, both for the speaker-dependent and the speaker-independent case. This effect holds across tasks (see Table~\ref{tab:pb-all-bin}) using the prosody feature set\footnote{We observe this on the Mel feature set as well.}. This may be due to the fact that the CNN looks for relative patterns in the data independent of their absolute position and values; and prosodic events are characterized by relative changes in speech. Normalizing the values may lead to a loss of fine differences in the data since the range of the values is decreased by z-scoring. The CNN performance in our experiments, however, appears to benefit from the original differences.

\section{Conclusion}
This paper presents experimental results using CNNs for word-based PER on low-level acoustic features, while emphasizing the effect of including context information. We show that the model performs well just by learning from simple frame-based features, and that the performance can be increased by adding position indicating features to the input that represents the word and its context.
Our model generalizes well from a speaker-dependent setup to a speaker-independent setting, yielding 86.3\% and 83.6\% accuracy, respectively, for pitch accent detection. Even in the more challenging task of classifying ToBI types, we obtain results across speakers that are comparable to previous related work, that is 69\% accuracy for pitch accents and 87.3\% for phrase boundaries. Futhermore, the presented method can be readily applied to other datasets.
Although a more detailed analysis is necessary to evaluate the performance on individual event types, we conclude that this method is quite suitable to the task, especially given its efficiency.


\bibliographystyle{IEEEtran.bst}

\bibliography{paslu}

\end{document}